\documentclass{article}
\usepackage[utf8]{inputenc}
\usepackage[margin=1in]{geometry}
\usepackage{appendix}
\usepackage{graphics}
\usepackage{graphicx}
\usepackage{url}
\usepackage[utf8]{inputenc} 
\usepackage[T1]{fontenc}    
\usepackage{hyperref}       
\usepackage{url}            
\usepackage{booktabs}       
\usepackage{amsfonts}       
\usepackage{nicefrac}       
\usepackage{microtype}      
\usepackage{amsmath}
\usepackage[ruled,vlined]{algorithm2e}
\usepackage{graphicx}
\usepackage{caption}
\usepackage{subcaption}
\usepackage{multirow}
\usepackage{array}

\usepackage{todonotes}
\usepackage{wrapfig}
\usepackage{soul}

\title{\bf A Weighted Solution to SVM   Actionability and Interpretability }

\author{
   Samuel Marc Denton  \\  \href{mailto:sam.denton@columbia.edu}{sam.denton@columbia.edu} 
   \and  
   Ansaf Salleb-Aouissi \\
   \href{mailto:ansaf@cs.columbia.edu}{ansaf@cs.columbia.edu} \\\\

 }

\date{{Department of Computer Science }\\
   Columbia University \\
   New York, NY 10025 \\}

\begin{document}

\maketitle

\begin{abstract}

Research in machine learning  has successfully developed algorithms to build accurate classification models. However, in many real-world applications, such as healthcare, customer satisfaction, and environment protection, we want to be able to use the models to decide what actions to take.

We investigate the concept of {\em actionability} in the context of Support Vector Machines. 
Actionability is as important as  {\em interpretability} or {\em explainability} of machine learning models, an ongoing and important research topic. Actionability is the task that gives us ways to act upon machine learning models and their predictions.  

 This paper finds a solution to the question of actionability on both linear and non-linear 
 SVM models. Additionally, we introduce a way to account for weighted actions that allow for more change in certain features than others. We propose a gradient descent solution on the linear, RBF, and polynomial kernels, and  
we test the effectiveness of our models on both synthetic and real datasets. We are also able to explore the model's interpretability through the lens of actionability.
\end{abstract}

\section{Introduction}

SVM models are one of the best performing predictive models in machine learning, as they can effectively separate data into differing labels while maximizing the margin of the decision function. 
Despite the progress achieved with other predictive methods such as deep learning, SVM models remain competitive and compelling in areas likes healthcare, where data will never reach 
the large volumes required to train deep neural networks models. 
While there is extensive research on SVMs, there is little to no information about finding ways to use the margin to change labels of new datapoints.
If we are given a  datapoint which, based on the decision function of the SVM, is most likely an \textit{undesirable} outcome (e.g., patient at risk of cardiovascular disease), or a datapoint for which we know the class (e.g, a sick patient), we would like to use the SVM boundary to move the datapoint to a \textit{desirable} outcome. Moving the datapoint can be complex and challenging depending on the features themselves as well as the shape of the SVM boundary. 
Having this functionality would be relevant to many applications such as medical care, customer satisfaction, and environmental protection. 

In this paper, we tackle the question of 
``What do I do with the prediction?''
or ``How do I use a prediction model to act upon an undesirable label?''.
Previous research (e.g., \cite{eloviciB03,liu00multilevel,Ras2002G,Ras2003IIS,trepos05}) has 
defined {\em actionability} as giving a recommendation of the \textit{easiest} way for a user to change an "unsatisfactory situation" based on a set of rules.
Unfortunately, 
while rule-based classifiers allow for interpretability of the feature decisions, they are not the best performing models. 
Additionally, previous papers focus on the framework of defining costs of an action in rule-based classification and how to maximize gain for a given cost. In the context of an SVM, this part of the problem will be slightly simpler as we can assume we are in a Euclidean space and thus we will frame our problem as a minimization of a weighted Euclidean distance. To our knowledge, we are the first to propose a general approach for actionability, in the context of support vector machines. 

We illustrate our approach with \textit{atherosclerosis}, a serious disease of the arteries characterized by the deposition of plaques of fatty material on their inner walls, causing serious complications like a reduced blood flow in the big arteries, which increases the risk for heart attack and stroke.
Atherosclerosis has no known cure, however, prevention can play an important role in reducing patient risk, by keeping under control the main risk factors: high cholesterol and triglycerides, tobacco consumption, diabetes, high BMI, high blood pressure, and lack of activity. The way these factors intertwine is unknown, and hoping to reduce all the factors at once is unrealistic for most patients. 
In atherosclerosis, as in many other health problems, one challenge  that a physician might face is to make important decisions about what should be done to cure a patient or reduce his risk of a specific disease. 
The purpose of actionability here is to suggest {\em actions} that are fine-grained modifications of some {\em actionable} risk factors  in order to decrease the risk of atherosclerosis in patients. 

We define an action as a change in the value of some feature (e.g., weight of a patient, or his tobacco consumption). Our approach focuses on a framework where a datapoint is framed as a vector of numerical features. We will then use the vector to predict an outcome based on the given SVM model. The goal will be to minimize the \textit{weighted distance} of the action that shifts the features to a point along the margin of the desired outcome. The weighted distance will be some combination of the distance travelled in each feature multiplied by a weight that represents the challenge of moving each feature, that is how  actionable each feature is (e.g., how hard it is to stop smoking). The \textit{weight vector} used to weight these distances is decided by the user. An example of weighting, say in the application of assessing the risk of atherosclerosis, may be that the amount of alcohol consumed (easier to change) has a lower weight while BMI (harder to change) has a higher weight. The action returned will then be the change in each actionable feature required to achieve the desired outcome. 

Through the lens of actionability, we also investigate  \textit{interpretability},  the ability to explain machine learning models in a form that is understandable to a human \cite{Doshi17}. By setting the weight vector to all 1's, we  use the recommended actions to rank-order the pertinent features 
in the model.

In Section \ref{related}, we describe the related work on actionability. In Section \ref{set up}, we describe the problem set up. In Section \ref{linear SVM}, we explore the linear SVM solution.
In Section \ref{non-linear SVM}, we extend the solution to fit the non-linear SVM problem. In Section \ref{Experiments}, we test the model on both simulated data and medical data. In Section \ref{conclusion}, we discuss the results and conclusions we can make about the algorithm. 

\section{Related Work}
\label{related}

In many real-world applications,  
a domain expert can act upon a prediction in many ways, but ultimately would like to use the prediction and the model  to decide which actions to take. 
This type of knowledge falls under the umbrella of actionability, an important but widely unexplored concept in machine learning. 
A related topic is the interpretability of machine learning models, for which there is a surge in research interest from the community (e.g., \cite{NIPS2017_6993,lei-etal-2016-rationalizing,lime2016}).  Understanding why a model is making a prediction 
increases our confidence in using the models, helps detect bias, and communicate/explain predictions to other  parties.
While interpretability aims to answer the question ``Why did the model make this prediction?'', actionability  
gives  ways to act upon those models and their predictions and address the related key question  of 
``What to do now?''. 

Most research on learning actionable knowledge goes back to rule-based models,   
that can be classified into three main lines of research: 
(1) use of predefined set of simple actions that are mapped to classification rules or deviations \cite{AdomaviciusT1997,EloviciSK03,JiangWTF05,Shapiro94}; (2)  sift through a set of association or subgroup discovery rules \cite{Gamberger02,Lavrac2002ICMLWorkshop,Lavrac02,Liu01}  to identify descriptive actionable rules
of the form $\text{Condition} \to \text{Class}$, where {\em Condition} is a conjunction of features describing the target {\em Class}; (3) elaborate actions by comparing pairs of classification rules with contradicting classes but the same values for all stable attributes \cite {Ras2002G,Ras2003IIS, Ras2005,Tsay2005} to generate action rules. 
Given an unsatisfactory situation, the work in \cite{trepos05,Trepos2011BuildingAF} discovers a set of action recommendations 
to improve that situation using classification rules.
Finally, actionability is formalized as a case-based reasoning problem in \cite{Yang2002,Yang2003}. Given an example to reclassify, the authors propose to search for the closest cases with the desirable class label, among training examples,  centroids of the clusters of the opposite class, or  
SVM support vectors. Choosing the closest support vector will be used as a baseline to compare our proposed approach. 
For a detailed literature survey on rule-based actionability see \cite{Trepos2011BuildingAF}.

\section{Presenting the SVM Actionability Problem}
\label{set up}

Our model learns how to move a datapoint from the undesirable class to the desirable class. We will provide our general framework for both a linear and non-linear SVM. For a given linear SVM model, we have a decision boundary given by $\mathbf{v}^T\mathbf{x} + b = 0$. Where $\mathbf{x} \in \mathbb{R}^d$ (the input space). We choose to use vector $\mathbf{v}$ here as the decision vector because we will later use $\mathbf{w}$ for the  weight vector used in the actionability problem. In the case of a non-linear SVM, the decision boundary is given by $\mathbf{v}^T\phi(\mathbf{x}) + b = 0$, where $\phi$ is the feature expansion used for the SVM.

Additionally, we are given an initial point $\mathbf{x_0}$ for which we predict an unfavorable result, or for which we know the label is not desirable (say with label $y_0$). Note, for this formulation, we require that for any label $y$, $y \in \{-1, +1\}$. We are also given a weight vector $\mathbf{w}$ which symbolizes the weight vector of the \textit{loss} of acting on a feature (ranging between $0$ and $1$). This means the higher $w_i$ is, the harder it is to change that feature (i.e. it is \textit{static}) and the lower $w_i$ is, the easier it is to change that feature (i.e. it is \textit{actionable}). We assume that all features are normalized between $-1$ and $1$ so that we can also standardize the weight vector and the distance formulation problem.

The goal will be to find the nearest point $\mathbf{x_n}$ on the other side of the SVM boundary (including the margin). In other words, for the case of a linear SVM, the output is shown in Equation \ref{Linear Output}.

\begin{equation}
    \label{Linear Output}
    \begin{gathered}
        min_{\mathbf{x_n} \in \mathbb{R}^d} dist(\mathbf{x_0}, \mathbf{x_n}) \\ \mbox{such that }\mathbf{v}^T\mathbf{x_n} + b + y_0 = 0
    \end{gathered}
\end{equation}

where $dist(\mathbf{x_0}, \mathbf{x_n})$ is defined in Equation \ref{Weighted Distance}. 
Equation \ref{non-linear Output} presents the case of a non-linear SVM.
\begin{equation}
    \label{non-linear Output}
    \begin{gathered}
        min_{\mathbf{x_n} \in \mathbb{R}^d} dist(\mathbf{x_0}, \mathbf{x_n}) \\ 
        \mbox{such that }\mathbf{v}^T\phi(\mathbf{x_n}) + b + y_0 = 0 \Longleftrightarrow \sum_{i=1}^{N} \alpha_i y_i K(\mathbf{x_i}, \mathbf{x_n}) + b + y_0 = 0 
    \end{gathered}
\end{equation}

where $dist(\mathbf{x_0}, \mathbf{x_n})$ is once again defined in Equation \ref{Weighted Distance}, $N$ is the number of training points, $\mathbf{x_i}$ is a training point, $\alpha_i$ is the associated dual coefficient of training point $\mathbf{x_i}$, $y_i$ is the label for training point $\mathbf{x_i}$, and $K$ is the kernel function corresponding to the feature expansion $\phi$.

The function we are aiming to minimize is the weighted Euclidean distance between $\mathbf{x_0}$ and $\mathbf{x_n}$ in the \textit{original} feature space ($\mathbb{R}^d$). This is shown in Equation \ref{Weighted Distance}.
\begin{equation}
    \label{Weighted Distance}
    dist(\mathbf{x_0}, \mathbf{x_n}) = \left(\mathbf{w} \odot \left[\mathbf{x_0} - \mathbf{x_n}\right]\right)^T\left(\mathbf{w} \odot \left[\mathbf{x_0} - \mathbf{x_n}\right]\right)
\end{equation}
where $\odot$ is the Hadamard product, or element-wise multiplication.

\paragraph{Static Features}
Throughout this paper, we use the weight vector $\mathbf{w}$ to account for differing levels of actionable vs. static features. It is important to note, by making incredibly uneven weight values where some features have weights around $0.001$ while static features have weights at $1$, they will for all intents and purposes be static. This is the most interesting case for this paper. However, it should be noted that if we had a static variable such as age that cannot be moved at all, we could easily fix that variable. To do this, we can factor the static features into the intercept variable $b$ for each point $\mathbf{x_0}$. Let's say $x_{0_0}$ and $x_{0_1}$ are static features. Then we can modify the original constraint as given in Equation \ref{Static features}.
\begin{equation}
    \label{Static features}
    \begin{gathered}
        \mathbf{v}^T\mathbf{x_n} + b + y_0 = 0 \Longleftrightarrow \mathbf{v'}^T\mathbf{x_n'} + v_0x_{0_0} + v_1x_{0_1} + b + y_0 = 0 \\
        \mathbf{v'}^T\mathbf{x_n'} + b' + y_0 = 0 \mbox{, such that } b' = b + v_0x_{0_0} + v_1x_{0_1}
    \end{gathered}
\end{equation}
In this case, we now have $\mathbf{v}', \mathbf{x_n}' \in \mathbb{R}^{d-2}$. We can then simply perform the original optimization problem with the features in a lower dimension leaving the static features as the fixed inputs of $\mathbf{x_0}$.

\begin{figure}[htbp]
  \begin{subfigure}[b]{0.5\textwidth}
    \includegraphics[width=\textwidth]{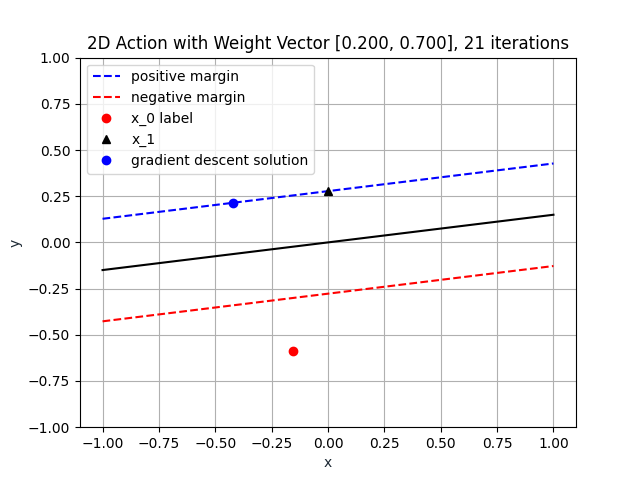}
    \caption{Changing negative label to positive label}
    \label{Example Linear SVM:1}
  \end{subfigure}
  \begin{subfigure}[b]{0.5\textwidth}
    \includegraphics[width=\textwidth]{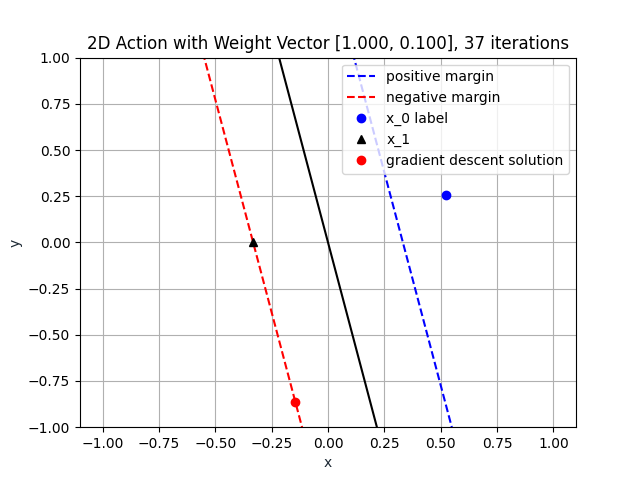}
    \caption{Changing positive label to negative label}
    \label{Example Linear SVM:2}
  \end{subfigure}
  \caption{Examples of changing the label predicted by a linear SVM. In Figure \ref{Example Linear SVM:1}, the weights are $w_x = 0.2$, $w_y = 0.7$. On the other hand, in Figure \ref{Example Linear SVM:2}, the weights are $w_x = 1.0$, $w_y = 0.1$.}

  \label{Example Linear SVM}

\end{figure}
\section{Solution to Linear SVM}
\label{linear SVM}

We derive a solution for Equation \ref{Linear Output} using a Lagrangian and then performing gradient descent to optimize for distance. This derivation is shown in Appendix \ref{Linear SVM Solution via Gradient Descent}. The result is to choose $\mathbf{x_1}$ such that $\mathbf{v}^T\mathbf{x_1} + b + y_0 = 0$ and perform iterative updates $\mathbf{x_{j+1}} := \mathbf{x_j} - \eta \nabla\mathcal{L}(\mathbf{x_j})$ where $\nabla_{\mathbf{x_j}}\mathcal{L}(\mathbf{x_j}, \lambda)$ is defined in Equation \ref{Linear Gradient}.
\begin{equation}
    \label{Linear Gradient}
    \nabla_{\mathbf{x_j}}\mathcal{L}(\mathbf{x_j}, \lambda) := -2\left(\mathbf{w} \odot \left[\mathbf{x_0} - \mathbf{x_j}\right]\right)  + \left(\frac{2\mathbf{v}^T\left(\mathbf{w} \odot \left[\mathbf{x_0} - \mathbf{x_j}\right]\right)}{\mathbf{v}^T \mathbf{v}}\right) \mathbf{v}
\end{equation}
In Figure \ref{Example Linear SVM} there are two examples showing the gradient descent solution for a linear SVM. Both solutions are heavily influenced by the given weights for $x$ and $y$ respectively.

\section{Solution to Non-Linear SVM}
\label{non-linear SVM}
Just as we did for the linear SVM, we can formulate the solution to Equation \ref{non-linear Output} using a Lagrangian and then perform gradient descent to optimize for the distance while taking into account the constraint. This derivation is shown in Appendix \ref{Non-Linear SVM Solution via Gradient Descent}. The result is to choose $\mathbf{x_1}$ such that $\sum_{i=1}^{N} \alpha_i y_i K(\mathbf{x_i}, \mathbf{x_1}) + b + y_0 = 0$. For some problems, finding $\mathbf{x_1}$ is computationally infeasible, so an approximation is needed. In our experiment we use the nearest support vector on the margin. Then use $\nabla_{\mathbf{x_j}}\mathcal{L}(\mathbf{x_j}, \lambda)$ defined in Equation \ref{Non-linear Gradient} in the gradient descent algorithm.
\begin{equation}
    \label{Non-linear Gradient}
    \begin{gathered}
        \nabla_{\mathbf{x_j}}\mathcal{L}(\mathbf{x_j}, \lambda) = -2\left(\mathbf{w} \odot \left[\mathbf{x_0} - \mathbf{x_j}\right]\right) + \lambda \left(\sum_{i=1}^{N} \alpha_i y_i K'(\mathbf{x_i}, \mathbf{x_n})\right),
        \\
        \lambda = \frac{2\left(\sum_{i=1}^{N} \alpha_i y_i K'(\mathbf{x_i}, \mathbf{x_n})\right)^T\left(\mathbf{w} \odot \left[\mathbf{x_0} - \mathbf{x_n}\right]\right)}{\left(\sum_{i=1}^{N} \alpha_i y_i K'(\mathbf{x_i}, \mathbf{x_n})\right)^T \left(\sum_{i=1}^{N} \alpha_i y_i K'(\mathbf{x_i}, \mathbf{x_n})\right)}
    \end{gathered}
\end{equation}

\begin{figure}[htbp]
 \begin{subfigure}[b]{0.5\textwidth}
  \centerline{ \includegraphics[scale=0.4]{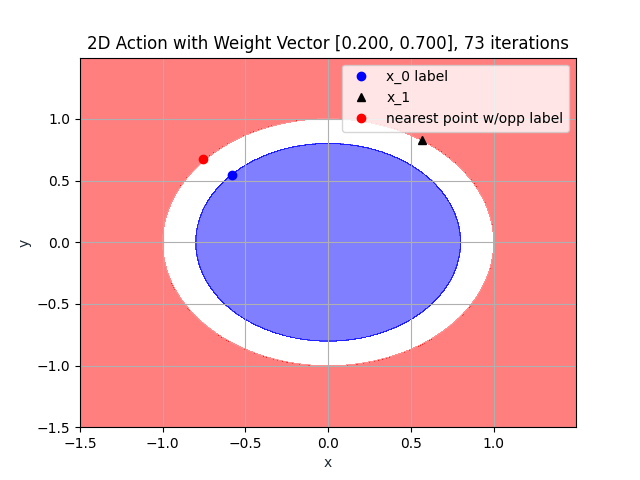}}
   \caption{Polynomial kernel example.}
   \label{Example_Circle_Poly}
  \end{subfigure}
  \begin{subfigure}[b]{0.5\textwidth}
  \centerline{\includegraphics[scale=0.4]{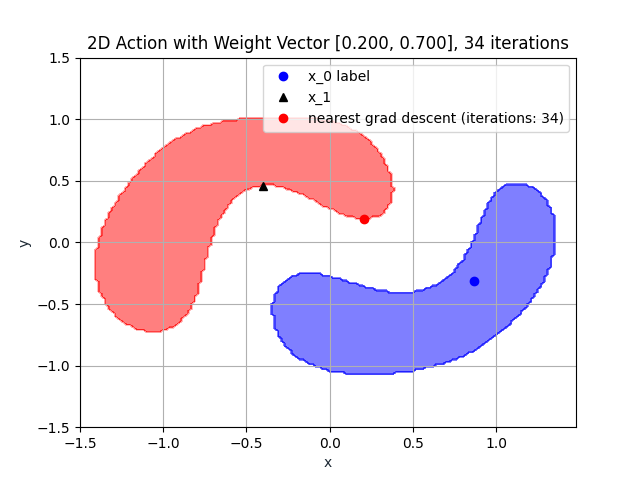}}
  \caption{RBF kernel example.}
  \label{Example_Moons_RBF}
  \end{subfigure}
  \caption{Examples with non-linear SVM}
  \label{Example of non linear SVM}
\end{figure}

where $K'(\mathbf{x_i}, \mathbf{x_n})$ is the derivative of the kernel function $K(\mathbf{x_i}, \mathbf{x_n})$ with respect to $\mathbf{x_n}$ (this must exist to find the nearest point).

Figure \ref{Example_Circle_Poly}, presents an example for a polynomial SVM that finds the nearest point with the opposite label to the initial point using gradient descent.

Figure \ref{Example_Moons_RBF} presents an example for a RBF SVM that finds the nearest point with the opposite label to the initial point using gradient descent.

\section{Experiments}
\label{Experiments}

We evaluate our approach and present our experiments on both simulated and medical data. 
We also introduce a method of implementing interpretability through the suggested actions.

\subsection{Simulated Data}
\label{simulated data}

We simulate data using the \href{https://scikit-learn.org/stable/datasets/index.html}{sklearn datasets package}. This package creates separable data for an SVM classifier with noise. We use three different simulated datasets: 1) linearly separable data 2) a dataset where one group is inside a circle and the other is outside (the circles dataset) and 3) a dataset of two interleaving half circles (the moons dataset). To measure the success of the action proposed by our solution, we compare it to the only other research that's been done in SVM actionability which is to move to the nearest support vector that's at least on the margin of the desired class \cite{Yang2002}.

\paragraph{Linear SVM Results}

\begin{figure}[htbp]
  \begin{subfigure}[b]{0.325\textwidth}
    \includegraphics[width=\textwidth]{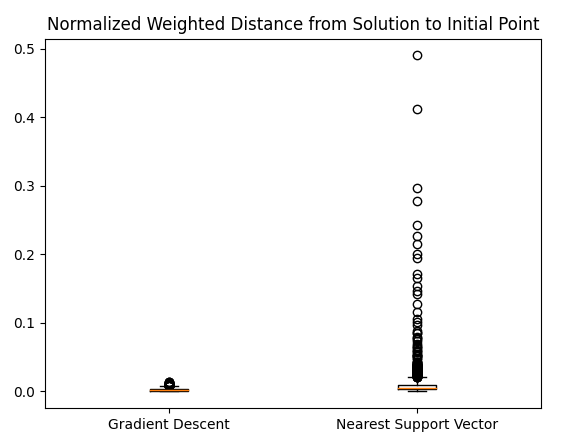}
    \caption{Linear SVM}
    \label{Simulated Distances:1}
  \end{subfigure}
  \begin{subfigure}[b]{0.335\textwidth}
    \includegraphics[width=\textwidth]{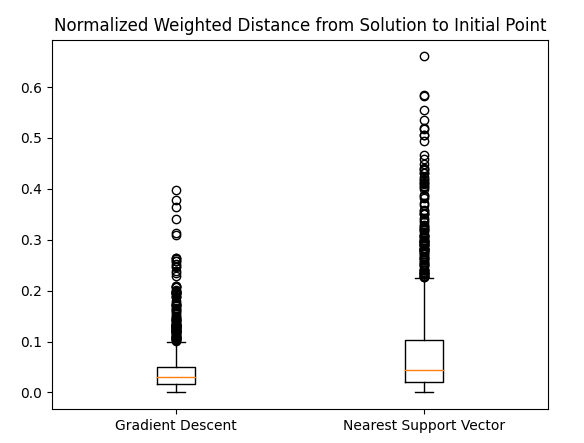}
    \caption{RBF SVM on moons}
    \label{Simulated Distances:2}
  \end{subfigure}
  \begin{subfigure}[b]{0.325\textwidth}
    \includegraphics[width=\textwidth]{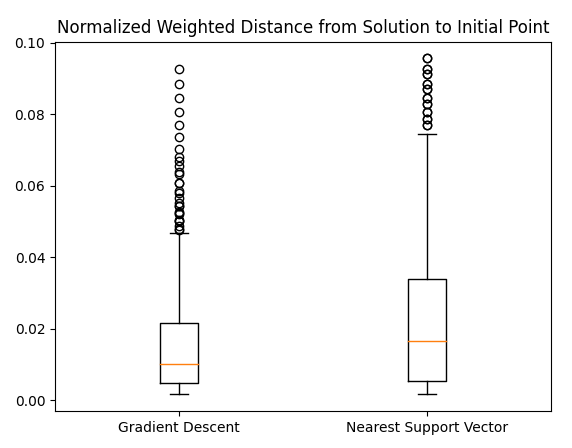}
    \caption{Polynomial SVM for circles}
    \label{Simulated Distances:3}
  \end{subfigure}
  \caption{Distances of the action for linear, polynomial and RBF SVM.}
  \label{Simulated Distances}
\end{figure}

In this section we compare the gradient descent and nearest support vector solutions. We see the distances for 1,000 datapoints in Figure \ref{Simulated Distances:1}. The average weighted normalized distance for the gradient descent solution was $0.003$ and for the nearest support vector was $0.013$. The gradient descent solution shows a statistically significant p-value $< 0.05$ which implies that the distance from gradient descent to the initial point is less than that of the nearest support vector. The gradient descent found a solution in at most $2.60$ milliseconds. 

\paragraph{Non-Linear SVM Results}

To compare the gradient descent and nearest support vector solutions, the distances for 1,000 datapoints with a RBF kernel are shown in Figure \ref{Simulated Distances:2} for the moons dataset.
The average weighted normalized distance for the gradient descent solution was $0.044$ and for the nearest support vector was $0.085$. The gradient descent solution shows a statistically significant p-value $< 0.05$ which implies that the distance from gradient descent to the initial point is less than that of the nearest support vector. The gradient descent found a solution in at most $916.72$ milliseconds.
The distances for 1,000 datapoints with a polynomial kernel are shown in Figure \ref{Simulated Distances:3} for the circles datasets. The average weighted normalized distance for the gradient descent solution was $0.015$ and for the nearest support vector was $0.022$. The gradient descent solution shows a statistically significant p-value $< 0.05$ which again implies that the distance from gradient descent to the initial point is less than that of the nearest support vector. The gradient descent found a solution in at most $64.6$ milliseconds.

\subsection{The Atherosclerosis dataset} \label{Atherosclerosis Data}

\begin{figure}[htbp]
  
\begin{center}
  \begin{subfigure}[b]{0.45\textwidth}
    \includegraphics[width=\textwidth]{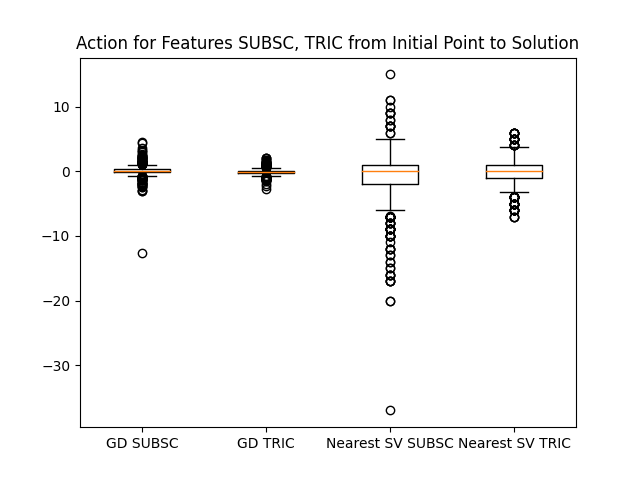}
    \caption{Action for RBF SVM}
    \label{Feature_Change:1}
  \end{subfigure}
  \begin{subfigure}[b]{0.45\textwidth}
    \includegraphics[width=\textwidth]{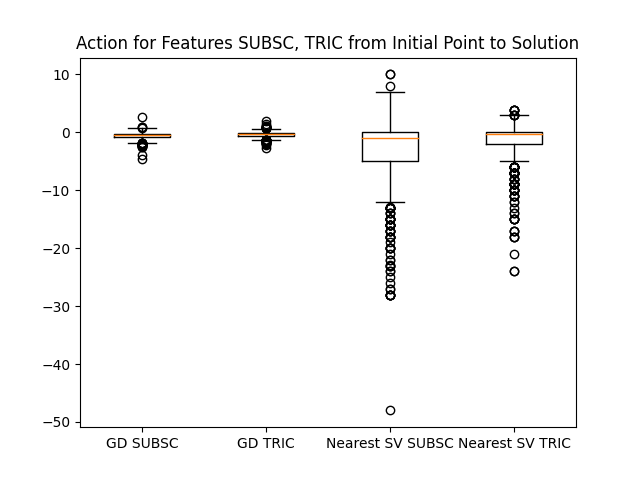}
    \caption{Action for polynomial SVM}
    \label{Feature_Change:2}
  \end{subfigure}
  \begin{subfigure}[b]{0.45\textwidth}
    \includegraphics[width=\textwidth]{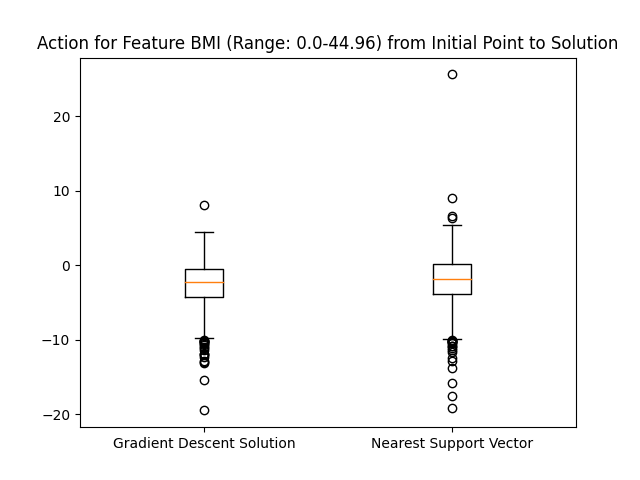}
    \caption{Action for RBF SVM}
    \label{Feature_Change:3}
  \end{subfigure}
  \begin{subfigure}[b]{0.45\textwidth}
    \includegraphics[width=\textwidth]{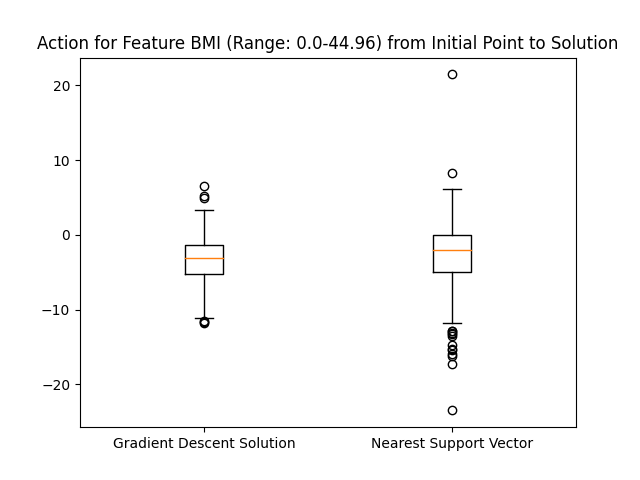}
    \caption{Action for polynomial SVM}
    \label{Feature_Change:4}
  \end{subfigure}
  \caption{Figures \ref{Feature_Change:1}-\ref{Feature_Change:2} are of SUBSC/TRIC action. Figures \ref{Feature_Change:3}-\ref{Feature_Change:4} are of BMI action. We compare gradient descent solution and nearest support vector. }
  \label{Feature_Change}
  \end{center}
\end{figure}

To illustrate the usefulness of our approach in preventing atherosclerosis, we conducted experiments using the atherosclerosis dataset derived from {\em Stulong}\footnote{\url{https://relational.fit.cvut.cz/dataset/Atherosclerosis}}, a twenty year study of the risk factors of the atherosclerosis in a population of 1,419 middle aged men.
The patients have been classified into three groups: the normal, the risk,  and the pathological groups. We also dispose of a granular integer risk score for each patient.  The dataset used was pre-processed for the  PKDD Discovery Challenge 2002. Our aim is to produce actions that would find the easiest way to take a patient at risk of atherosclerosis and get them to a point that would have a prediction of no risk. 

When creating a SVM model to describe this dataset, we used the following features  [SUBSC, TRIC, TRIGL, SYST, DIAST, BMI, WEIGHT, CHLST, ALCO\_CONS, TOBA\_CONSO] as these have all shown impacts on atherosclerosis risk. There is a description of each of these features in Appendix \ref{Description of Atherosclerosis Risk Factors}. We then grouped together the pathological risk and risk groups as \textbf{positive risk} or $+1$ and no risk as \textbf{negative risk} or $-1$. Then, we tuned the SVM to perform best on this prediction problem using cross validation and tested on a test set of 20\% of the data. This process was performed 10 times. The RBF SVM achieved an f1-score of $0.68$  on $540$ negative risk examples and an f1-score of $0.91$ on $1960$ positive risk examples with an overall accuracy of $0.86$. On the other hand, the polynomial SVM achieved an f1-score of $0.61$ on $540$ negative risk examples and an f1-score of $0.91$ on $1960$ positive risk examples with an overall accuracy of $0.85$. This process led us to the best parameters for the two models: an RBF SVM with hyperparameters $\gamma = 1, C = 10$ and a polynomial SVM with hyperparameter degree $= 4$.

With the SVM models defined, the next step was to test the recommended actions based on the solution presented in Section \ref{non-linear SVM}. For each point that has a label of $+1$ and has an SVM prediction of $+1$, we found two points: one, the \textbf{nearest support vector} with a label of $-1$ and two, the solution via \textbf{gradient descent} of the nearest point on the margin presented in Section \ref{non-linear SVM}. Note, for this example, we choose a weight vector of $[1, 1, .5, .5, .5, .2, .2, .1, .05, .05]$ to correspond with the features [SUBSC, TRIC, TRIGL, SYST, DIAST, BMI, WEIGHT, CHLST, ALCO\_CONS, TOBA\_CONSO]. For the RBF SVM, the algorithm took a maximum time of $3.99s$ and for the polynomial SVM, the algorithm took a maximum time of $1.25s$.

We compare these two solutions using three different methods: 
(1) measure the weighted distance from both solutions to the point we are moving, (2)  measure the action recommended by both solutions, and (3)  measure the predicted risk of both solutions. To measure (3), we create a linear model to predict the granular integer risk score of atherosclerosis which is measured from $1 - 10$. This linear model is presented in Equation \ref{Linear Model}. Using this regression model\footnote{Interestingly, observe the negative coefficient associated with alcohol consumption. It has been noted in previous studies that some alcohol consumption can be beneficial on cardiovascular health \cite{Lucas2002}}, we can calculate and compare the risk associated with the old and new datapoints. This will help assess the impact of the actions in reducing the risk of atherosclerosis in the patient.

{ \small  \begin{multline} \label{Linear Model}
    Risk = 0.0072 \cdot SUBSC + 0.0036 \cdot TRIC + 0.00004 \cdot TRIGL + 0.0088 \cdot SYST \\
     + 0.0068 \cdot DIAST + 0.0191 \cdot BMI + 0.0037 \cdot WEIGHT + 0.0049 \cdot CHLST \\
     - 0.1903 \cdot ALCO\_CONS + 0.8468 \cdot TOBA\_CONS - 1.0558
\end{multline}
}

\vspace*{-0.4cm}

\begin{wrapfigure}{r}{0.38\textwidth}
\vspace*{-0.5cm}
\includegraphics[scale=0.4]{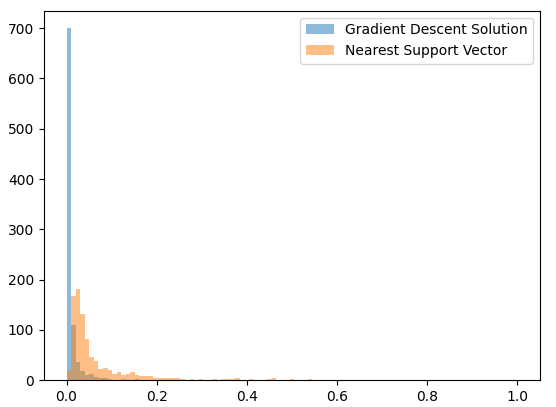}
\vspace*{-0.2cm}
\caption{RBF Distances Histogram}
\label{RBF_Distances_Histogram}
\vspace*{-0.4cm}
\end{wrapfigure}

In Figures \ref{Feature_Change:1}-\ref{Feature_Change:2}, we see a box plot for the actions for a RBF SVM and a polynomial SVM for two of the highly weighted features: SUBSC and TRIC. As we would hope, the gradient descent solution requires less movement than the nearest support vector for these features that are extremely hard to change. On the other hand, when investigating a feature with a lower weight, such as BMI, there is more relative movement, particularly for the gradient descent solution, as this is a feature that a patient can more easily change. This result is shown in Figures \ref{Feature_Change:3}-\ref{Feature_Change:4}. The rest of these feature action boxplots are shown in Appendix \ref{Atherosclerosis Feature Boxplots}.

For the RBF SVM, the average weighted distance of the gradient descent solution was $0.011$ and for the nearest support vector was $0.073$. For the polynomial SVM, the average weighted distance of the gradient descent solution was $0.012$ and for the nearest support vector was $0.144$. The maximum time to run the gradient descent algorithm on either kernel was $2.265s$. For both the RBF and polynomial SVM, our solution is statistically significantly closer to the original point than the solution presented by the nearest support vector. These results are presented as a histogram for the RBF SVM in Figure \ref{RBF_Distances_Histogram}. The same plot is shown for the polynomial SVM in Appendix \ref{Atherosclerosis Polynomial Histogram}. Additionally, we see these results presented in the form of a boxplot for both the RBF and polynomial SVM in Appendix \ref{Atherosclerosis Distances Boxplots}. 

\begin{figure}[ht]
\begin{center}
  \begin{subfigure}[b]{\textwidth}
    \centerline{\includegraphics[scale=0.6]{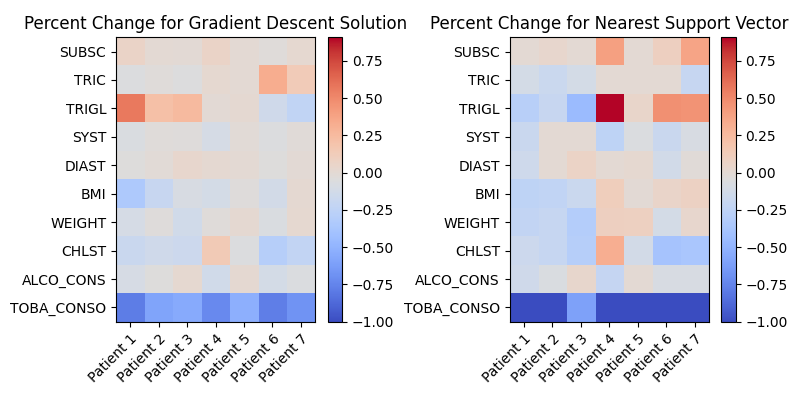}}
    \caption{Action for RBF SVM}
    \label{Example_Patients:1}
  \end{subfigure}
  
  \begin{subfigure}[b]{\textwidth}
    \centerline{\includegraphics[scale=0.6]{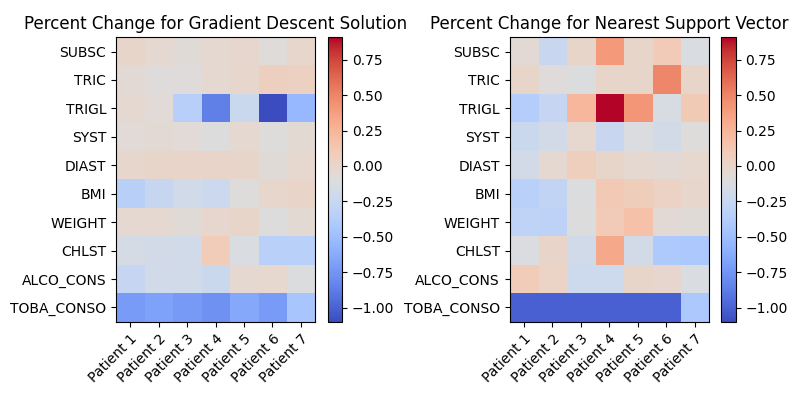}}
    \caption{Action for Poly SVM}
    \label{Example_Patients:2}
  \end{subfigure}
  \caption{Depicting Actions for 7 Patients using both RBF and Poly SVMs.}
  \label{Example_Patients}
  \end{center}
\end{figure}

\begin{figure}[ht]
\begin{center}
  \begin{subfigure}[b]{0.45\textwidth}
    \includegraphics[width=\textwidth]{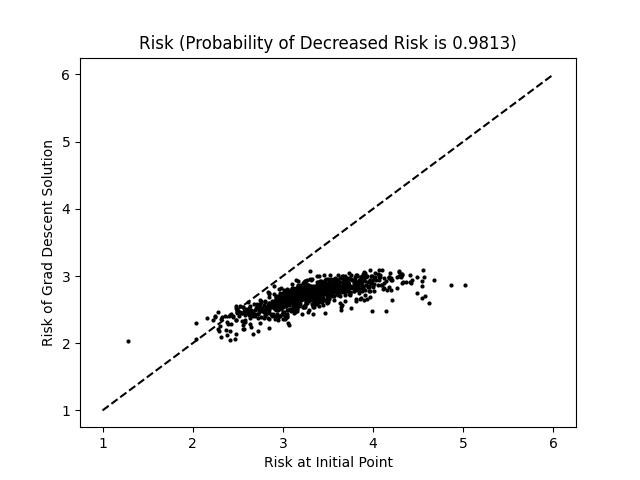}
    \label{Decreased Risk RBF:1}
  \end{subfigure}
  \begin{subfigure}[b]{0.45\textwidth}
    \includegraphics[width=\textwidth]{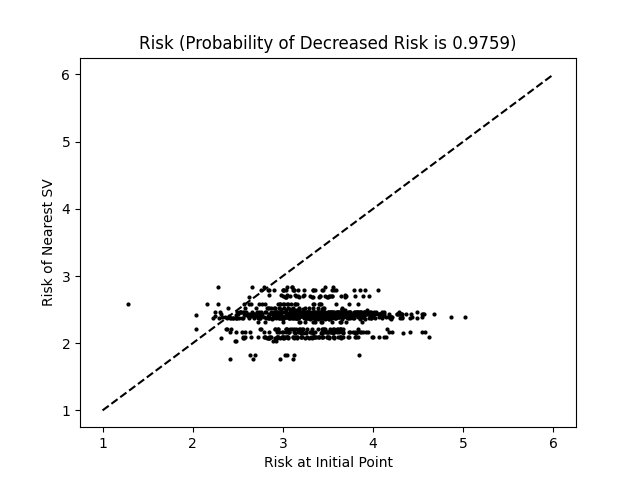}
    \label{Decreased Risk RBF:2}
  \end{subfigure}
  \caption{Comparing the predicted risk of the initial point against the predicted risk of the suggested point of both algorithms for a RBF SVM.}
  \label{Decreased Risk RBF}
\end{center}
\end{figure}

In Figure \ref{Decreased Risk RBF}, we see a graph of the predicted risk of both the original point and the suggested point for both the gradient descent solution and the nearest support vector for the RBF SVM. The gradient descent solution shows a higher probability of decreasing the risk of the initial point. For the polynomial SVM, the same conclusion is found which is that the gradient descent solution shows a higher probability of decreasing the risk of the initial point. The figure depicting these results is shown in Appendix \ref{Atherosclerosis Decreased Risk Poly}. 

Finally, we have included a set of figures to visualize the recommended actions of both the gradient descent solution and the nearest support vector. In Figure \ref{Example_Patients:1}, we visualize the percent change in each feature recommended for the solutions using a RBF SVM. In Figure \ref{Example_Patients:2}, we see the same for a polynomial SVM. Although these are just a few randomly chosen patients, we see the desired trend that the changes required for the gradient descent solutions are fainter colors (smaller percent change) while the changes for the nearest support vector are darker colors (larger percent change).

\subsubsection{Atherosclerosis Interpretability}
\label{Atherosclerosis Interpretability Section}

Using the framework presented in the processing of the atherosclerosis dataset, we present a way to delve into the interpretability of the SVM that classifies patients as either negative or positive risk. To do this, we use results of the SVM actionability.

\begin{table}[htbp]
{
    \begin{center}
    \begin{tabular}{|m{3.5cm}||m{4.5cm}|m{5cm}|}
    \hline
     & Normalized Amplitude Mean & Normalized Amplitude Median \\
    \hline
    TOBA\_CONS & 0.1989 & 0.1689\\
    CHLST & 0.0750 & 0.0604\\
    BMI & 0.0749 & 0.0666\\
    SYST & 0.0661 & 0.0508\\
    TRIGL & 0.0552 & 0.0444\\
    DIAST & 0.0525 & 0.0398\\
    TRIC & 0.0407 & 0.0308\\
    WEIGHT & 0.0376 & 0.0265\\
    SUBSC & 0.0373 & 0.0269\\
    ALCO\_CONS & 0.0294 & 0.0195\\
    \hline
    \end{tabular}
    \caption{Feature interpretability} 
    \label{Atherosclerosis Interpretability}
    \end{center}
    }
\end{table}

By using the same RBF SVM ($\gamma = 1, C = 10$) and a weight vector of all $1$s, we can investigate the suggested actions as a proxy for feature importance. That is, when all features have equal weight, the features that move most to switch a classification from risk to no risk are the most important features when decreasing atherosclerosis risk. We use this weight vector of all $1$s and measure the amplitude of all of the recommended actions (normalized by feature) to determine which features required the most change. In Table \ref{Atherosclerosis Interpretability}, we present the median and mean of the normalized amplitudes of the suggestion actions of all the features. Based on the results, changing TOBA\_CONS is the most important (by a substantial amount) in terms of decreasing the risk while ALCO\_CONS is the least important. The rest of the rankings allow one to identify the features an at-risk patient should focus most on when attempting to reduce risk of atherosclerosis. Polynomial SVM results are in Appendix \ref{Polynomial Atherosclerosis Interpretability}.

\section{Discussion and Conclusion}
\label{conclusion}
The task of devising actions or actionability based on predictive models has been relatively unexplored in modern machine learning research.
We propose a simple, yet an effective and efficient way of using SVM models for actionability. 
We show an answer to the nearest point of any point that switches the label predicted by a linear SVM to a desired result. This provides a good solution to any application that implements a linear SVM to divide data. 
For a non-linear SVM, we were able to find an algorithm via gradient descent that predicts the nearest point with a desirable outcome. This reduces the cost of the action. 
However, it is important to note that this algorithm 
is sensitive to the initial guess for the nearest point. A search of possible starting points should allow one to pick a point that will lend itself to finding the true global optima (such as the nearest support vector). 
In Section \ref{Atherosclerosis Data}, we saw that the gradient descent solution gave an easier action than the nearest support vector with the given weight vector for both the RBF and polynomial SVM. Additionally, the gradient descent solution gave a higher probability of reducing predicted risk than the nearest support vector. 
{Using the recommended actions, we explored the interpretability of the SVM models for the atherosclerosis dataset, and concluded  that tobacco consumption was the most important risk factor to adjusting the atherosclerosis risk.}

We must note that the action is only as good as the SVM model itself. If the SVM model is a good fit for the prediction problem, then the resulting action will also be relevant. 
Additionally, to reduce uncertainty of a given action, one could adjust the algorithm to go \textit{beyond} the positive/negative margin to reach a point that is more confidently classified in the desired group. 
In future work, it would be interesting to explore variations of the actionability problem. For example, it's possible that a user has the bandwidth to change a limited number of features.
Additionally, it would not be hard to expand the above results to a multiclass SVM classifier. We could then provide the nearest action to the desired class. 
Finally, there is  more work to be done in exploring the weighting. Are there other more applicable ways to define actionable and static features? What other metrics could we use to replace the distance/cost metric that we aim to minimize? 
Paired with visualization techniques, our SVM approach to actionability has the potential to have an immense impact in assisting physicians in 
preventing diseases with actionable risk factors. 
We hope that this paper will motivate the machine learning community to conduct research on  actionability so to take full advantage of the accurate prediction models. 

\section{Broader Impact}
\label{broader impact}

This algorithm is an extension of the SVM algorithm, so therefore it will impact the community in similar ways that SVM has impacted the world. A suggested action on a predicted outcome of the SVM algorithm could have a potentially huge beneficial impact in {\bf healthcare} and the  world of {\bf precision medicine}. 
Say a patient is told they will most likely need surgery, a doctor could then explain to the patient the easiest changes they could make so that the doctor would no longer recommend surgery as a solution. Thus, actionability could be used to decide on a prevention policy by providing  the physician with a tool for investigating the effects of the various factors on the risk of disease, for an individual patient. 
Additionally, this could be used for a positive impact in the world of {\bf climate change}. It's possible that researchers use SVMs to predict outcomes of certain environments and that actionability would allow them to suggest how communities could achieve more desirable outcomes.

On the other hand, SVMs can be used
by companies who may be using SVM models to predict the success of an advertisement. This algorithm could hypothetically used in {\bf marketing and business strategies} to help companies attract more customers which may be viewed negatively in the eyes of the consumer. Similarly, politicians may be able to use this to change opinions of voters by feeding certain features that tend to change the minds of voters for their own gain. 

While SVMs and the actionability approach we propose can be used maliciously, we do not believe this research would inherently put anyone at a disadvantage. However, we recognize that there are a few faults of this system. There is inherent bias in the weights provided by the user and therefore users could manipulate the system to give recommendations that enforce a previous bias. In some ways, it may be better to use this as an unweighted algorithm or require that extensive research is done into the weights of the actions.

\bibliographystyle{plain} 
\bibliography{references,jmlr}

\newpage

\appendix

\section{Linear SVM Solution via Gradient Descent}
\label{Linear SVM Solution via Gradient Descent}
We can formulate the solution to Equation \ref{Linear Output} using a Lagrangian and then perform gradient descent to optimize for the distance while taking into account the constraint. The Lagrangian is formulated in Equation \ref{Linear Lagrangian}. We then take the gradient of $\mathcal{L}(\mathbf{x_n}, \lambda)$ with respect to $\mathbf{x_n}$.
\begin{equation}
    \label{Linear Lagrangian}
    \begin{gathered}
        \mathcal{L}(\mathbf{x_n}, \lambda) = \left(\mathbf{w} \odot \left[\mathbf{x_0} - \mathbf{x_n}\right]\right)^T\left(\mathbf{w} \odot \left[\mathbf{x_0} - \mathbf{x_n}\right]\right) + \lambda \left(\mathbf{v}^T\mathbf{x_n} + b + y_0\right) \\ 
        \nabla_{\mathbf{x_n}}\mathcal{L}(\mathbf{x_n}, \lambda) = -2\left(\mathbf{w} \odot \left[\mathbf{x_0} - \mathbf{x_n}\right]\right) + \lambda \mathbf{v}
    \end{gathered}
\end{equation}
Since $\mathcal{L}(\mathbf{x_n}, \lambda)$ is convex, it will be optimized when the gradient is equal to zero which allows us to solve for $\lambda$. To do this, take the dot product of the gradient with the vector $\mathbf{v}$ which leads to the result in Equation \ref{Linear Lambda}.
\begin{equation}
    \label{Linear Lambda}
    \begin{gathered}
        -2\left(\mathbf{w} \odot \left[\mathbf{x_0} - \mathbf{x_n}\right]\right) + \lambda \mathbf{v} = 0 \\
        -2\mathbf{v}^T\left(\mathbf{w} \odot \left[\mathbf{x_0} - \mathbf{x_n}\right]\right) + \lambda \mathbf{v}^T \mathbf{v} = 0
        \\
        \lambda = \frac{2\mathbf{v}^T\left(\mathbf{w} \odot \left[\mathbf{x_0} - \mathbf{x_n}\right]\right)}{\mathbf{v}^T \mathbf{v}}
    \end{gathered}
\end{equation}
We can then use this equation for $\lambda$ to find the $\lambda$ at any $\mathbf{x_j}$. This gives us a value for $\nabla_{\mathbf{x_j}}\mathcal{L}(\mathbf{x_j}, \lambda)$ at any vector $\mathbf{x_j}$. Algorithm \ref{Linear SVM Grad Descent} shows the gradient descent method for finding $\mathbf{x_n}$ where $\nabla_{\mathbf{x_j}}\mathcal{L}(\mathbf{x_j}, \lambda) := -2\left(\mathbf{w} \odot \left[\mathbf{x_0} - \mathbf{x_j}\right]\right)  + \left(\frac{2\mathbf{v}^T\left(\mathbf{w} \odot \left[\mathbf{x_0} - \mathbf{x_j}\right]\right)}{\mathbf{v}^T \mathbf{v}}\right) \mathbf{v}$.

\begin{algorithm}[ht]
\SetAlgoLined
\caption{Gradient Descent for Solving Actionability on Linear SVM}
 \textbf{Input:} SVM decision vector $\mathbf{v}$, bias term $b$, initial point $\mathbf{x_0}$ with label $y_0$, and weight vector $\mathbf{w}$ \;
 set stopping tolerance $\epsilon > 0$\;
 set $\mathbf{x_1}$ such that $\mathbf{v}^T\mathbf{x_1} + b + y_0 = 0$\;
 $\mathbf{x_j} = \mathbf{x_1}$\;
 \While{$\left[\mathbf{x_0} - \mathbf{x_j}\right]^T\left[\mathbf{x_0} - \mathbf{x_j}\right] > \epsilon$}{
  $\mathbf{x_{j+1}} := \mathbf{x_j} - \eta \nabla\mathcal{L}(\mathbf{x_j}) $\;
  $\mathbf{x_{j}} = \mathbf{x_{j+1}}$;
 }
 $\mathbf{x_n} = \mathbf{x_j}$\;
 \KwResult{The vector $\mathbf{x_n}$ that minimizes the weighted distance to $\mathbf{x_0}$}
 \label{Linear SVM Grad Descent}
\end{algorithm}

\section{Non-Linear SVM Solution via Gradient Descent}
\label{Non-Linear SVM Solution via Gradient Descent}
Just as we did for the Linear SVM, we can formulate this question using a Lagrangian and then perform gradient descent to optimize for the distance while taking into account the constraint. The Lagrangian is formulated in Equation \ref{non-linear Lagrangian}. We then take the gradient of $\mathcal{L}(\mathbf{x_n}, \lambda)$ with respect to $\mathbf{x_n}$.
\begin{equation}
    \label{non-linear Lagrangian}
    \begin{gathered}
        \mathcal{L}(\mathbf{x_n}, \lambda) = \left(\mathbf{w} \odot \left[\mathbf{x_0} - \mathbf{x_n}\right]\right)^T\left(\mathbf{w} \odot \left[\mathbf{x_0} - \mathbf{x_n}\right]\right) + \lambda \left(\sum_{i=1}^{N} \alpha_i y_i K(\mathbf{x_i}, \mathbf{x_n}) + b + y_0\right) \\ 
        \nabla_{\mathbf{x_n}}\mathcal{L}(\mathbf{x_n}, \lambda) = -2\left(\mathbf{w} \odot \left[\mathbf{x_0} - \mathbf{x_n}\right]\right) + \lambda\left(\sum_{i=1}^{N} \alpha_i y_i K'(\mathbf{x_i}, \mathbf{x_n})\right)
    \end{gathered}
\end{equation}
In Equation \ref{non-linear Lagrangian}, $K'(\mathbf{x_i}, \mathbf{x_n})$ is the derivative of the kernel function $K(\mathbf{x_i}, \mathbf{x_n})$ with respect to $\mathbf{x_n}$ (this must exist to find the nearest point). Again, like we did in the Linear SVM, we will set the gradient equal to 0 and then solve for $\lambda$. To do this, take the dot product of the equation with $\left(\sum_{i=1}^{N} \alpha_i y_i K'(\mathbf{x_i}, \mathbf{x_n})\right)$ which leads to the result in Equation \ref{non-linear Lambda}.
\begin{equation}
    \label{non-linear Lambda}
    \begin{gathered}
        -2\left(\mathbf{w} \odot \left[\mathbf{x_0} - \mathbf{x_n}\right]\right) + \lambda \left(\sum_{i=1}^{N} \alpha_i y_i K'(\mathbf{x_i}, \mathbf{x_n})\right) = 0 \\
        -2\left(\sum_{i=1}^{N} \alpha_i y_i K'(\mathbf{x_i}, \mathbf{x_n})\right)^T\left(\mathbf{w} \odot \left[\mathbf{x_0} - \mathbf{x_n}\right]\right) + \lambda \left(\sum_{i=1}^{N} \alpha_i y_i K'(\mathbf{x_i}, \mathbf{x_n})\right)^T \left(\sum_{i=1}^{N} \alpha_i y_i K'(\mathbf{x_i}, \mathbf{x_n})\right) = 0 \\
        \lambda = \frac{2\left(\sum_{i=1}^{N} \alpha_i y_i K'(\mathbf{x_i}, \mathbf{x_n})\right)^T\left(\mathbf{w} \odot \left[\mathbf{x_0} - \mathbf{x_n}\right]\right)}{\left(\sum_{i=1}^{N} \alpha_i y_i K'(\mathbf{x_i}, \mathbf{x_n})\right)^T \left(\sum_{i=1}^{N} \alpha_i y_i K'(\mathbf{x_i}, \mathbf{x_n})\right)}
    \end{gathered}
\end{equation}
We can then use the equation for $\lambda$ from Equation \ref{non-linear Lambda} when minimizing the distance for any $\mathbf{x_j}$ (simply replace $\mathbf{x_n}$ with the current estimate $\mathbf{x_j}$). This gives us a real value for $\nabla_{\mathbf{x_j}}\mathcal{L}(\mathbf{x_j}, \lambda)$ at any vector $\mathbf{x_j}$. Using $\lambda$ defined in Equation \ref{non-linear Lambda} we define $\nabla_{\mathbf{x_j}}\mathcal{L}(\mathbf{x_j}, \lambda)$ in Equation \ref{non-linear Gradient Equation}.
\begin{equation}
    \label{non-linear Gradient Equation}
    \nabla_{\mathbf{x_j}}\mathcal{L}(\mathbf{x_j}, \lambda) = -2\left(\mathbf{w} \odot \left[\mathbf{x_0} - \mathbf{x_j}\right]\right) + \lambda \left(\sum_{i=1}^{N} \alpha_i y_i K'(\mathbf{x_i}, \mathbf{x_n})\right)
\end{equation}
The gradient descent algorithm for a non-linear SVM is the same as Algorithm \ref{Linear SVM Grad Descent}, except that we must choose $\mathbf{x_1}$ such that $\sum_{i=1}^{N} \alpha_i y_i K(\mathbf{x_i}, \mathbf{x_1}) + b + y_0 = 0$. For some problems finding $\mathbf{x_1}$ is computationally infeasible, so an approximation is needed. In our experiment we use the nearest support vector on the margin. Then use $\nabla_{\mathbf{x_j}}\mathcal{L}(\mathbf{x_j}, \lambda)$ defined in Equation \ref{non-linear Gradient Equation} in the gradient descent algorithm.

\subsection{Specific Kernel Derivatives}
To use the gradient descent solution for a kernel, we must define $K'(\mathbf{x}, \mathbf{y})$. The two kernels we explore in this paper are the \textbf{RBF kernel} defined as $K(\mathbf{x}, \mathbf{y}) := exp(-\gamma ||\mathbf{x} - \mathbf{y}||^2)$ and the \textbf{polynomial kernel} defined as $K(\mathbf{x}, \mathbf{y}) := \left(\mathbf{x}^T \mathbf{y} + c\right)^D$. 

\textbf{RBF Kernel:} the derivative for the RBF kernel (with respect to $\mathbf{y}$) is defined as 
$$K'(\mathbf{x}, \mathbf{y}) := 2\gamma(\mathbf{x} - \mathbf{y}) K(\mathbf{x}, \mathbf{y}) = 2\gamma(\mathbf{x} - \mathbf{y})exp(-\gamma ||\mathbf{x} - \mathbf{y}||^2)$$

\textbf{Polynomial Kernel:} the derivative for the polynomial kernel (with respect to $\mathbf{y}$) is defined as 
$$K'(\mathbf{x}, \mathbf{y}) := D\mathbf{x}\left(\mathbf{x}^T \mathbf{y} + c\right)^{D-1}$$

\section{Description of Atherosclerosis Risk Factors}
\label{Description of Atherosclerosis Risk Factors}

\begin{table}[h]
{\footnotesize 
\begin{center}
\begin{tabular}{ |m{3cm}||m{6cm}|m{2.5cm}|}
\hline
 & Description & Feature Range \\
\hline
SUBSC & Skin fold subscapularis (mm)  & 4 - 70\\
TRIC & Skin fold triceps (mm) & 1 - 35\\
TRIGL & Triglycerides in mg\%  & 42 - 1197\\
SYST & Blood pressure systolic & 80 - 200\\
DIAST & Blood pressure diastolic & 50 - 125\\
BMI & Body Mass Index & 16.98 - 44.96\\
WEIGHT & Weight (kg) & 52 - 133\\
CHLST & Cholesterol in mg\% & 134 - 510\\
ALCO\_CONS & Alcohol consumption & 1 - 1.67\\
TOBA\_CONS & Tobacco consumption & 0 - 1.25\\
\hline
\end{tabular}
\end{center}
}
\caption{Description of the risk factors used for atherosclerosis SVM.}
\label{Avg. Suggestions}
\end{table}

\section{Histogram of Atherosclerosis Distances}
\label{Atherosclerosis Polynomial Histogram}
\begin{figure}[ht]
\begin{center}
        \includegraphics[scale=0.65]{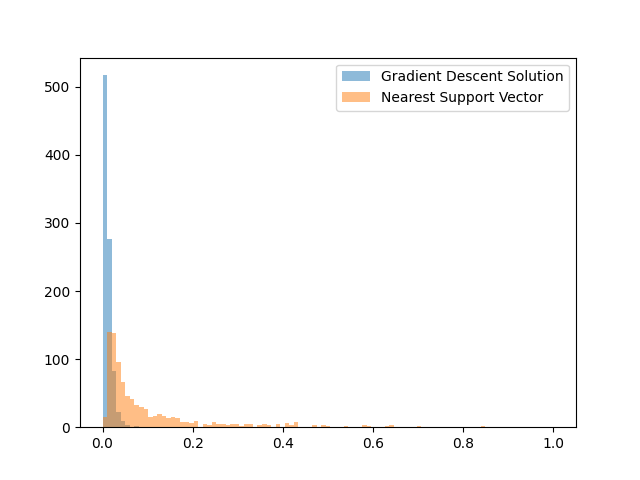}
    \label{Poly_Histogram:1}
    \caption{Distances for polynomial kernel}
  \label{Poly_Histogram}
\end{center}
\end{figure}

In Figure \ref{Poly_Histogram}, we see the histogram representation of the distances of the gradient descent solution and nearest support vector for a polynomial kernel.

\section{Boxplots of Atherosclerosis Distances}
\label{Atherosclerosis Distances Boxplots}
\begin{figure}[htbp]
  \begin{subfigure}[b]{0.5\textwidth}
    \includegraphics[width=\textwidth]{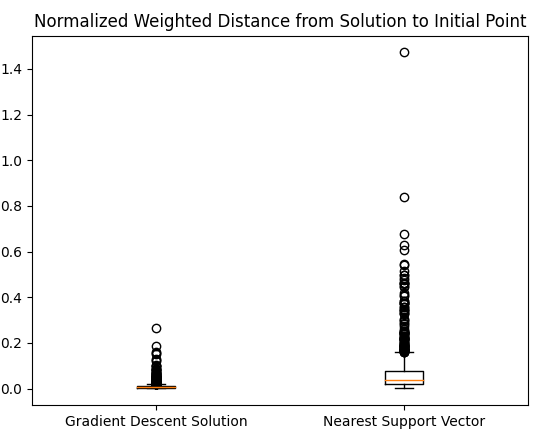}
    \label{Distance_Boxplot:1}
    \caption{Distances for RBF kernel}
  \end{subfigure}
  \begin{subfigure}[b]{0.5\textwidth}
    \includegraphics[width=1.05\textwidth]{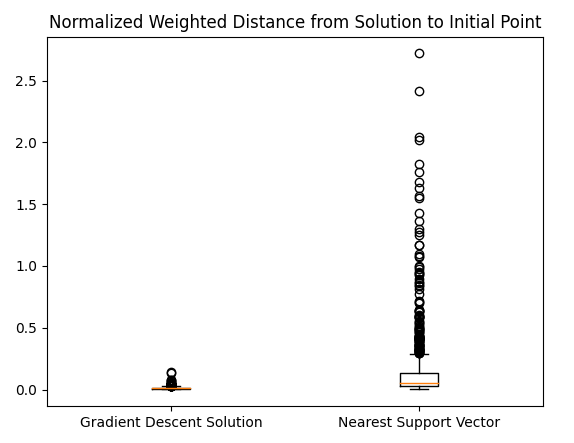}
    \label{Distance_Boxplot:2}
    \caption{Distances for polynomial kernel}
  \end{subfigure}
  \caption{Boxplot of distances for gradient descent solution and nearest support vector.}
  \label{Distance_Boxplot}
\end{figure}

In Figure \ref{Distance_Boxplot}, we see the distance comparisons between the gradient descent solution and nearest support vector.

\section{Boxplots of Feature Changes}
\label{Atherosclerosis Feature Boxplots}
\begin{figure}[htbp]
  \begin{subfigure}[b]{0.5\textwidth}
    \includegraphics[width=\textwidth]{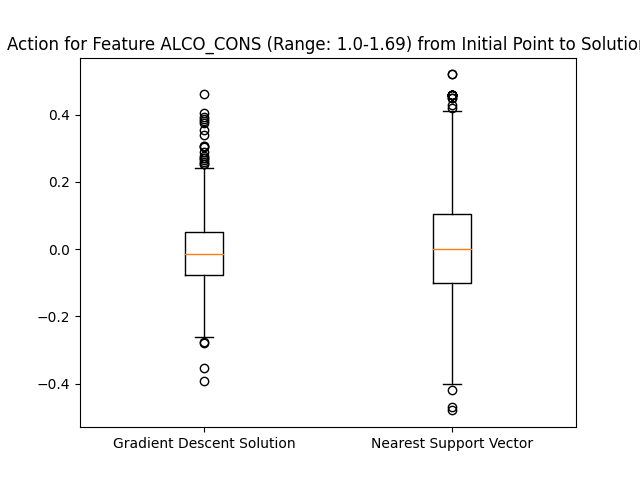}
    \label{ALCO_CONS:1}
    \caption{Action for RBF kernel}
  \end{subfigure}
  \begin{subfigure}[b]{0.5\textwidth}
    \includegraphics[width=\textwidth]{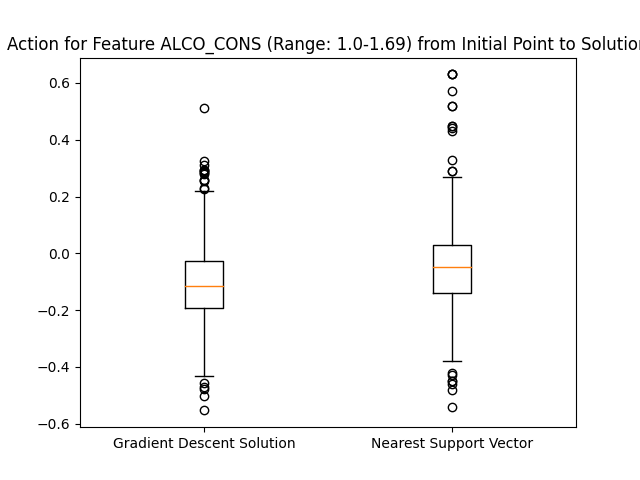}
    \label{ALCO_CONS:2}
    \caption{Action for polynomial kernel}
  \end{subfigure}
  \caption{Boxplot of ALCO\_CONS action for gradient descent solution and nearest support vector.}
  \label{ALCO_CONS}
\end{figure}

\begin{figure}[htbp]
  \begin{subfigure}[b]{0.5\textwidth}
    \includegraphics[width=\textwidth]{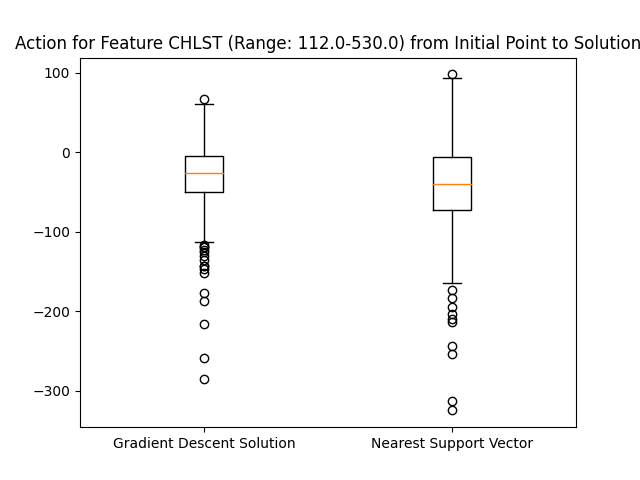}
    \label{CHLST:1}
    \caption{Action for RBF kernel}
  \end{subfigure}
  \begin{subfigure}[b]{0.5\textwidth}
    \includegraphics[width=\textwidth]{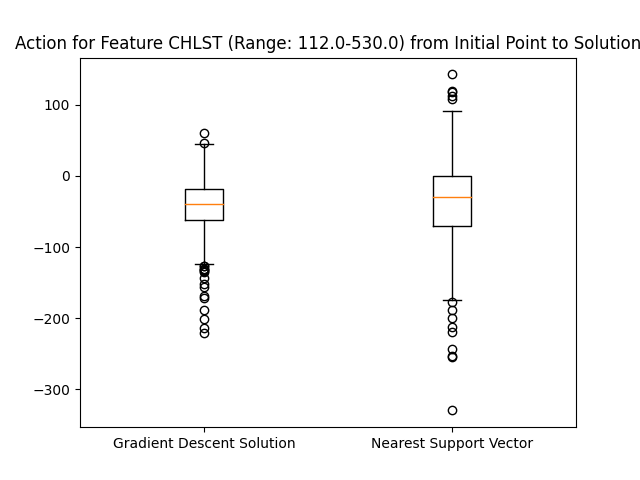}
    \label{CHLST:2}
    \caption{Action for polynomial kernel}
  \end{subfigure}
  \caption{Boxplot of CHLST action for gradient descent solution and nearest support vector.}
  \label{CHLST}
\end{figure}

\begin{figure}[htbp]
  \begin{subfigure}[b]{0.5\textwidth}
    \includegraphics[width=\textwidth]{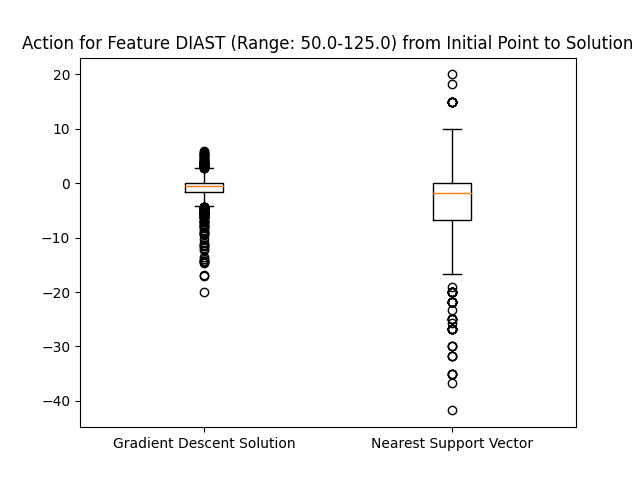}
    \label{DIAST:1}
    \caption{Action for RBF kernel}
  \end{subfigure}
  \begin{subfigure}[b]{0.5\textwidth}
    \includegraphics[width=\textwidth]{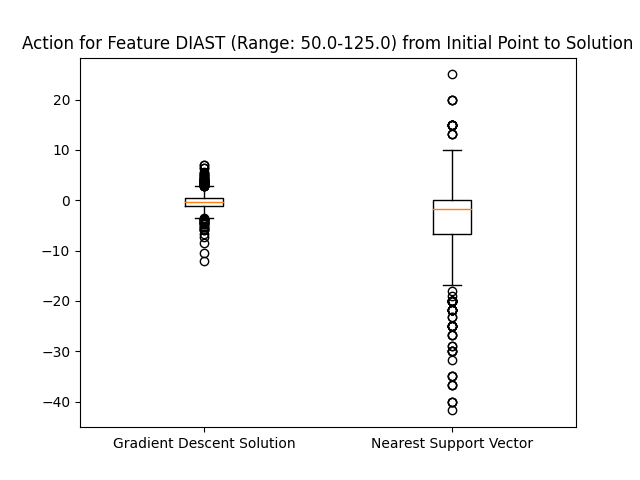}
    \label{DIAST:2}
    \caption{Action for polynomial kernel}
  \end{subfigure}
  \caption{Boxplot of DIAST action for gradient descent solution and nearest support vector.}
  \label{DIAST}
\end{figure}

\begin{figure}[htbp]
  \begin{subfigure}[b]{0.5\textwidth}
    \includegraphics[width=\textwidth]{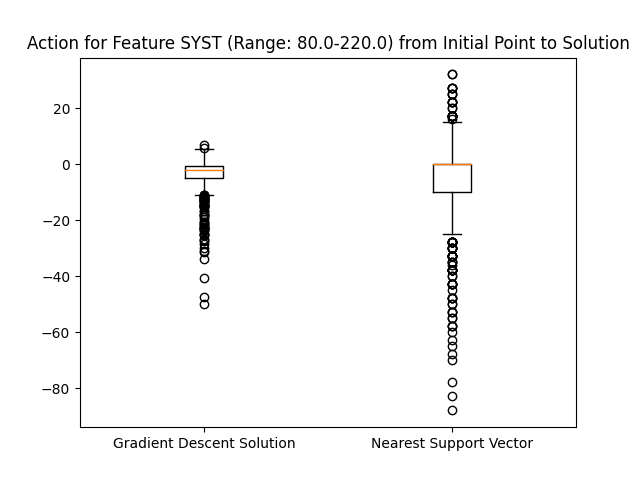}
    \label{SYST:1}
    \caption{Action for RBF kernel}
  \end{subfigure}
  \begin{subfigure}[b]{0.5\textwidth}
    \includegraphics[width=\textwidth]{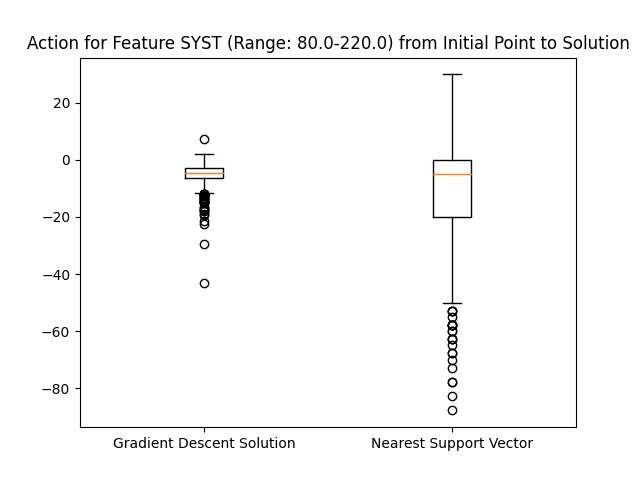}
    \label{SYST:2}
    \caption{Action for polynomial kernel}
  \end{subfigure}
  \caption{Boxplot of SYST action for gradient descent solution and nearest support vector.}
  \label{SYST}
\end{figure}

\begin{figure}[htbp]
  \begin{subfigure}[b]{0.5\textwidth}
    \includegraphics[width=\textwidth]{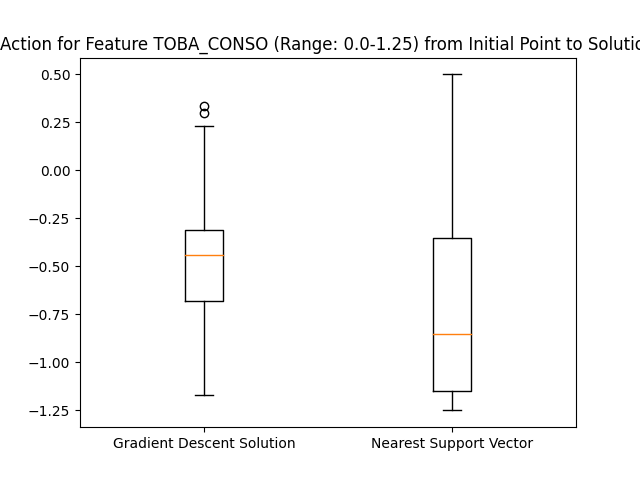}
    \label{TOBA_CONSO:1}
    \caption{Action for RBF kernel}
  \end{subfigure}
  \begin{subfigure}[b]{0.5\textwidth}
    \includegraphics[width=\textwidth]{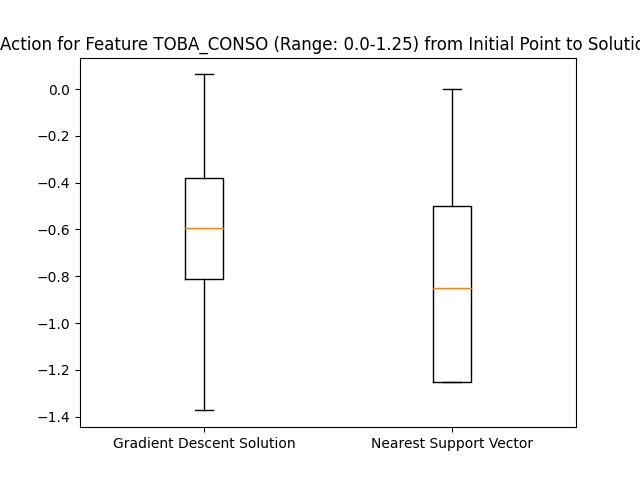}
    \label{TOBA_CONSO:2}
    \caption{Action for polynomial kernel}
  \end{subfigure}
  \caption{Boxplot of TOBA\_CONS action for gradient descent solution and nearest support vector.}
  \label{TOBA_CONSO}
\end{figure}

\begin{figure}[htbp]
  \begin{subfigure}[b]{0.5\textwidth}
    \includegraphics[width=\textwidth]{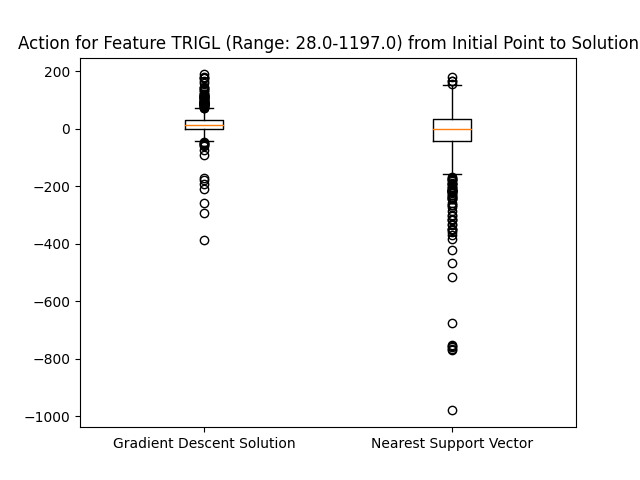}
    \label{TRIGL:1}
    \caption{Action for RBF kernel}
  \end{subfigure}
  \begin{subfigure}[b]{0.5\textwidth}
    \includegraphics[width=\textwidth]{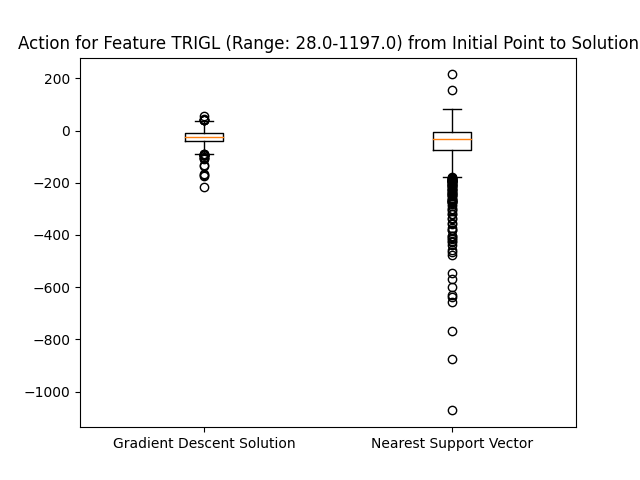}
    \label{TRIGL:2}
    \caption{Action for polynomial kernel}
  \end{subfigure}
  \caption{Boxplot of TRIGL action for gradient descent solution and nearest support vector.}
  \label{TRIGL}
\end{figure}

\begin{figure}[htbp]
  \begin{subfigure}[b]{0.5\textwidth}
    \includegraphics[width=\textwidth]{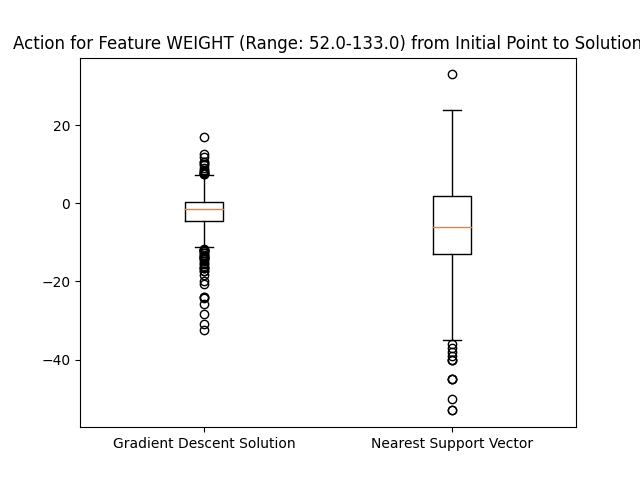}
    \label{WEIGHT:1}
    \caption{Action for RBF kernel}
  \end{subfigure}
  \begin{subfigure}[b]{0.5\textwidth}
    \includegraphics[width=\textwidth]{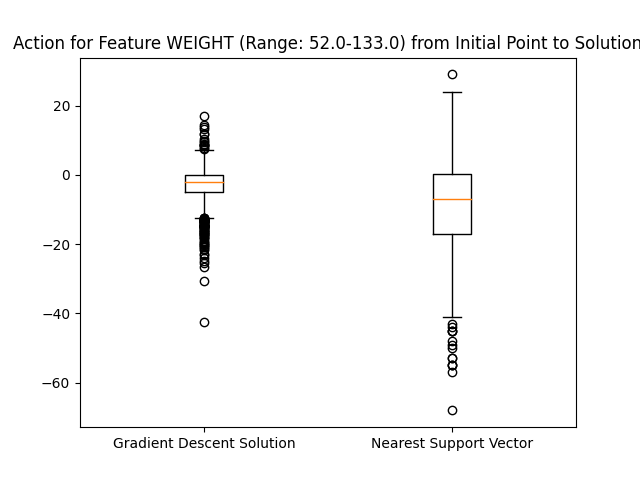}
    \label{WEIGHT:2}
    \caption{Action for polynomial kernel}
  \end{subfigure}
  \caption{Boxplot of WEIGHT action for gradient descent solution and nearest support vector.}
  \label{WEIGHT}
\end{figure}

In Figures \ref{ALCO_CONS} through \ref{WEIGHT}, we present the boxplots of the other feature changes not presented in the text for a RBF and polynomial SVM.

\section{Probability of Decreased Risk for Polynomial SVM}
\label{Atherosclerosis Decreased Risk Poly}

\begin{figure}[htbp]
  \begin{subfigure}[b]{0.5\textwidth}
    \includegraphics[width=\textwidth]{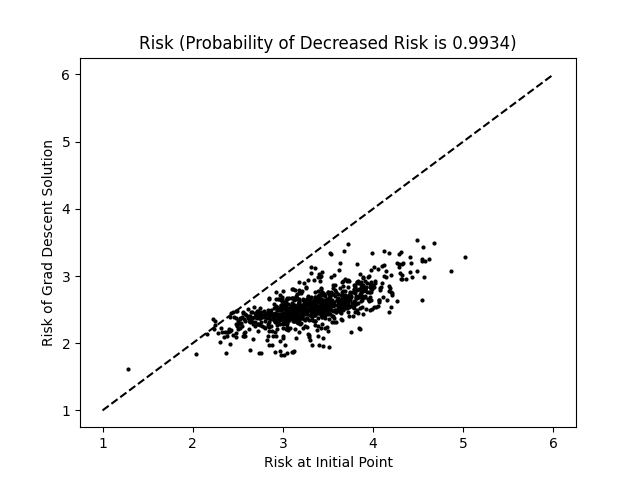}
    \label{Decreased Risk Poly:1}
  \end{subfigure}
  \begin{subfigure}[b]{0.5\textwidth}
    \includegraphics[width=\textwidth]{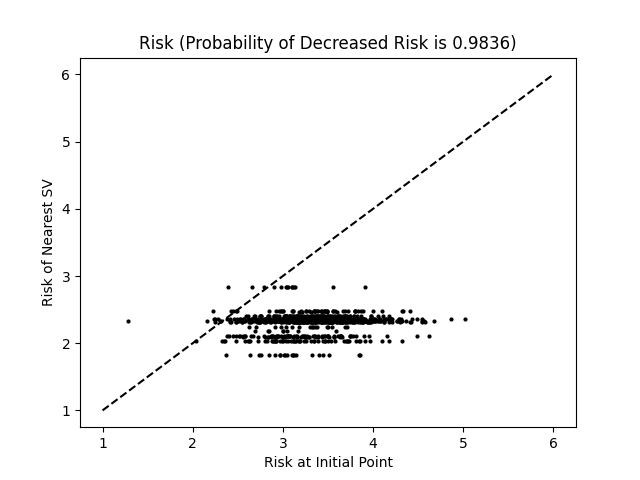}
    \label{Decreased Risk Poly:2}
  \end{subfigure}
  \caption{Comparing the predicted risk of the initial point against the predicted risk of the suggested point of both algorithms for a polynomial SVM. In both graphs, we see the probability of decreasing the predicted risk by moving from the initial point to the suggested point.}
  \label{Decreased Risk Poly}
\end{figure}

In Figure \ref{Decreased Risk Poly} we present the predicted risk of the initial point vs. the solution for a polynomial SVM.

\section{Polynomial SVM Atherosclerosis Interpretability}
\label{Polynomial Atherosclerosis Interpretability}

\begin{table}
    \begin{center}
    \begin{tabular}{|m{3.5cm}||m{4.5cm}|m{4.7cm}|}
    \hline
     & Normalized Amplitude Mean & Normalized Amplitude Median \\
    \hline
    TOBA\_CONSO & 0.2059 & 0.2035\\
    CHLST & 0.1159 & 0.1071\\
    TRIC & 0.1056 & 0.0744\\
    TRIGL & 0.0992 & 0.0854\\
    WEIGHT & 0.0900 & 0.0648\\
    SYST & 0.0834 & 0.0757\\
    ALCO\_CONS & 0.0785 & 0.0624\\
    BMI & 0.0707 & 0.0630\\
    DIAST & 0.0684 & 0.0576\\
    SUBSC & 0.0654 & 0.0536\\
    \hline
    \end{tabular}
    \caption{Feature interpretability for polynomial kernel (sorted by mean)}
    \label{Atherosclerosis Poly Interpretability}
    \end{center}
\end{table}

By using the same polynomial SVM (with $degree = 4$) and a weight vector of all $1$s, we investigate the suggested actions as a proxy for feature importance. In Table \ref{Atherosclerosis Poly Interpretability}, we present the median and mean of the normalized amplitudes of the suggestion actions of all the features. Based on the results, changing TOBA\_CONS is the most important (by a substantial amount) in terms of decreasing risk of atherosclerosis while SUBSC is the least important. The rest of the rankings are pretty similar to the rankings for the RBF SVM presented in Table \ref{Atherosclerosis Interpretability}.

\end{document}